\pdfoutput=1

\documentclass[11pt]{article}

\usepackage[]{EMNLP2023} 
\usepackage{graphicx}
\usepackage{times}
\usepackage{latexsym}
\usepackage[english]{babel}
\usepackage{amsmath}
\usepackage{hyperref}

\usepackage{amsthm}
\usepackage{algorithm}
\usepackage{algpseudocode}
\usepackage{graphicx}
\usepackage{subcaption}
\usepackage{amssymb}

\makeatletter
\title{Unified Locational Differential Privacy Framework}
\author{Aman Priyanshu\thanks{* Equal contribution} \\ \textbf{Yash Maurya}\footnotemark[1] \\\textbf{Suriya Ganesh}\footnotemark[1] \\\textbf{Vy Tran}\footnotemark[1] \\ \\School of Computer Science \\  Carnegie Mellon University}
\makeatother
\begin{document}

\maketitle

\begin{abstract}
Aggregating statistics over geographical regions is important for many applications, such as analyzing income, election results, and disease spread. However, the sensitive nature of this data necessitates strong privacy protections to safeguard individuals. In this work, we present a unified locational differential privacy (DP) framework to enable private aggregation of various data types, including one-hot encoded, boolean, float, and integer arrays, over geographical regions. Our framework employs local DP mechanisms such as randomized response, the exponential mechanism, and the Gaussian mechanism. We evaluate our approach on four datasets representing significant location data aggregation scenarios. Results demonstrate the utility of our framework in providing formal DP guarantees while enabling geographical data analysis.
\end{abstract}

\section{Introduction}

The aggregation of statistics over geographical regions is critical for a wide range of applications, from analyzing income inequality and assessing the effectiveness of public health interventions to allocating resources and understanding electoral trends. For instance, during the COVID-19 pandemic, aggregated data by ZIP code was instrumental in evaluating the success of quarantine measures and guiding the distribution of medical supplies. Similarly, aggregated income statistics are essential for the US Census Bureau to study economic disparities and inform policy decisions. In the political sphere, aggregated voting patterns by precinct or constituency are crucial for understanding election outcomes and identifying areas for targeted campaigning.

However, the data involved in these analyses, such as health records, financial information, and voting histories, is highly sensitive and requires robust privacy safeguards. Failing to protect individual privacy can lead to severe consequences, ranging from personal embarrassment and social stigma to financial harm and political persecution. Therefore, it is paramount to develop techniques that allow for the aggregation and analysis of such data while preserving the privacy of individuals.

Inspired by Apple's pioneering work on learning "iconic scenes" of popular locations using differential privacy \cite{DpApple}, we develop a unified DP framework for aggregating various types of geographical data while providing formal privacy guarantees. Apple's approach combines locally-perturbed vectors using secure aggregation to achieve DP, ensuring that individual contributions are masked while still allowing for meaningful learning. Building upon this foundation, our framework extends to handle a diverse range of data types commonly encountered in geographical analysis, including one-hot encoded vectors, boolean arrays, integer counts, and floating-point values. By accommodating these various data types, our framework enables private aggregation for a broad spectrum of use cases, from contagion tracking and income analysis to preference modeling and beyond.

We implement several local DP mechanisms, including randomized response, the exponential mechanism, and the Gaussian mechanism, using the diffprivlib library \cite{diffprivlib}. Finally, we use the Opacus library \cite{opacus} to track privacy budgets and ensure that the overall DP guarantees are maintained throughout the aggregation process.

The key objectives of this work was to:
\begin{enumerate}
    \item Present a unified DP framework that can handle diverse data types commonly found in geographical analysis, providing a versatile tool for private aggregation across a wide range of applications. 
    \item Evaluate our framework on carefully constructed datasets that represent significant location data aggregation scenarios, demonstrating its effectiveness and robustness in practice.
\end{enumerate}

With these objectives, we hope to contribute to the growing body of work on differentially private location analysis and provide a practical framework for researchers and practitioners alike.

\section{Background}
\subsection{Tools \& Libraries}

\subsubsection{Mechanism Implementation Tools}
Our locational DP framework heavily relies on the Diffprivlib \cite{diffprivlib} library for implementing the core DP mechanisms. Diffprivlib is a comprehensive Python library developed by researchers at IBM that provides a wide range of DP algorithms and tools. The library is designed to be user-friendly and flexible, making it well-suited for both research and practical applications. We employ Diffprivlib in our framework to implement three key local DP mechanisms: randomized response, the exponential mechanism, and the Gaussian mechanism. By leveraging its framework we were able to quickly perform our local DP computations efficiently. Also, the library's modular design also allows for easy integration of additional DP mechanisms, available within the library, in the future, enabling our framework to adapt to evolving privacy requirements and aggregation scenarios.

\subsubsection{Privacy Accounting Libraries}

To maintain DP guarantees throughout the data aggregation process, our framework employs Meta AI's Opacus \cite{opacus} library for privacy budget tracking and composition. Opacus is a powerful Python library that enables differentially private computations in PyTorch, with a focus on machine learning applications. However, its privacy accounting capabilities make it well-suited for our locational DP framework as well. One of the key challenges in implementing DP is keeping track of the privacy budget, which represents the total amount of privacy loss incurred by a series of DP computations. Opacus provides advanced privacy accounting techniques that allow for accurate tracking of the privacy budget across multiple mechanisms and aggregation steps. Also, its implementation of composition theorems, such as basic composition and advanced variants like Rényi DP, allows us to accurately account for the privacy loss at each stage of the aggregation process. By integrating Opacus into our framework, we ensure that the overall DP guarantees are maintained and that the privacy budget is not exceeded.

\subsection{Datasets}

\begin{figure*}
    \centering
    \includegraphics[width=0.75\textwidth]{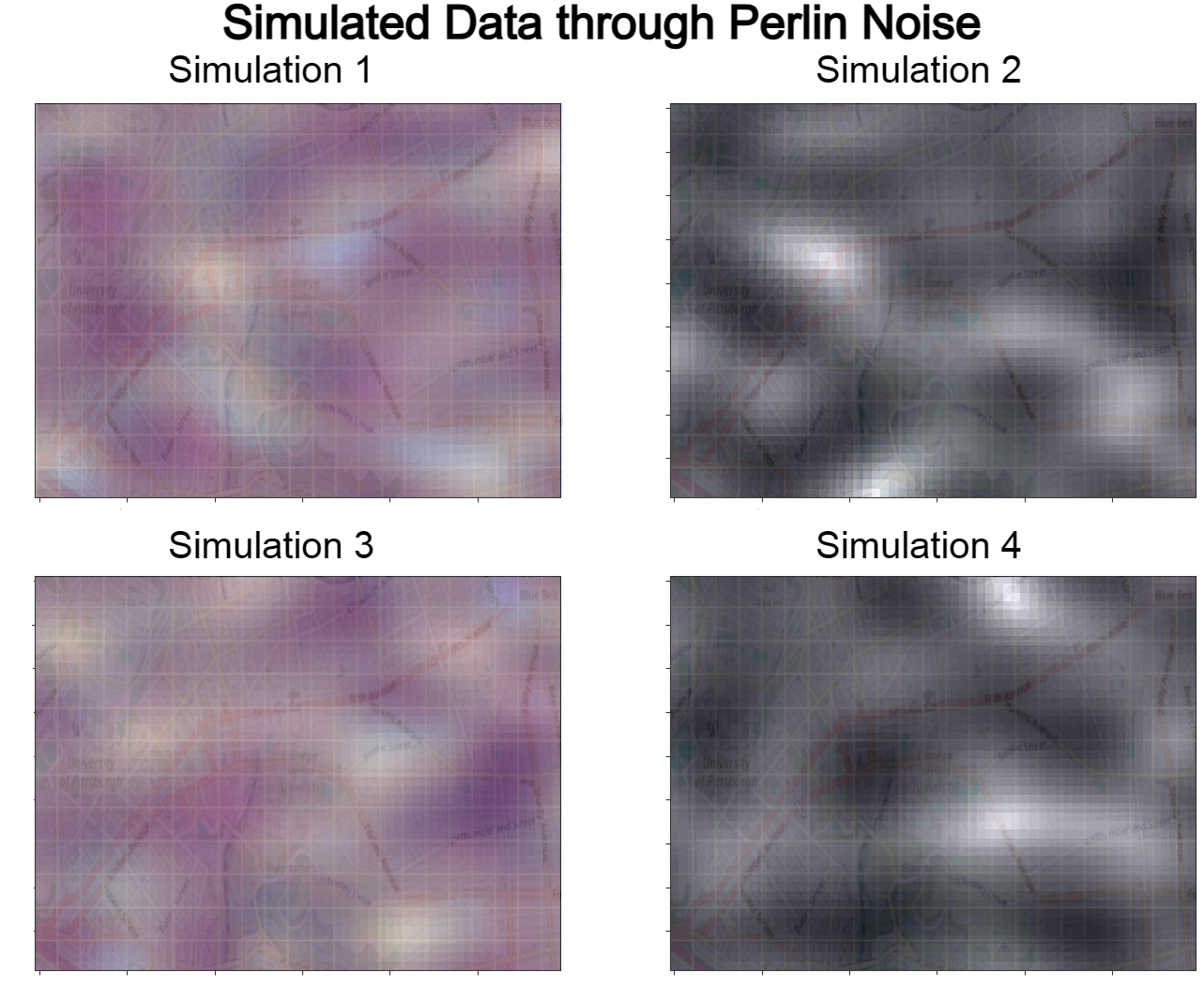}
  \caption{We present an abstraction of what our simulated data is describing.}
  \label{fig:perlin}
\end{figure*}

Due to the lack of suitable location-centric, real datasets that validate our framework, we simulate four datasets representing different data aggregation scenarios commonly encountered in real-world applications. These datasets are carefully designed to capture the key characteristics and challenges associated with various types of geographical data.

\begin{enumerate}
    \item \textbf{One-hot-encoded Data (Apple's Implementation)}: To validate our framework against Apple's approach, we focus on the Pittsburgh area and generate one-hot encoded data using overlapping Perlin noise and randomization with normalization. The one-hot encoding represents the presence or absence of specific features or attributes across different geographical locations. By employing overlapping Perlin noise, we introduce realistic spatial correlations among the features, mimicking the patterns observed in real-world data. Randomization and normalization ensure that the generated data falls within expected ranges and follows a realistic distribution.
    \item \textbf{Boolean-based Data (Contagion Tracking)}: Simulating contagion tracking or movement monitoring during pandemics like COVID-19, we generate boolean data using Perlin noise and randomization techniques. The boolean values represent the presence or absence of a certain condition or event at each geographical location. Perlin noise is used to create spatial clusters and hotspots, resembling the spreading patterns of contagions. Randomization introduces variability and noise into the data, reflecting the stochastic nature of real-world scenarios.
    \item \textbf{Integer-based Data (Rankings)}: To simulate scenarios involving ranking or preference data, such as election results or candidate rankings, we generate data using Perlin noise, randomization, and normalization. The generated values represent the rankings or scores assigned to different entities across geographical locations. Perlin noise is employed to create spatial patterns and dependencies among the rankings, capturing the notion of regional preferences or voting trends. Randomization and normalization ensure that the generated rankings are realistic and fall within expected ranges.
    \item \textbf{Float-based Data (Income)}: Representing scenarios like income bracket analysis or US Census data aggregation, we generate integer data using Perlin noise, randomization, and normalization techniques. The integer values correspond to different income brackets or categories associated with each geographical location. Perlin noise is used to introduce spatial correlations and dependencies among the income levels, reflecting the socio-economic patterns observed in real-world data. Randomization and normalization ensure that the generated data follows a realistic distribution and falls within expected ranges for each income bracket.
\end{enumerate}

Across all datasets, we employ Perlin noise to introduce realistic spatial correlations, and randomization with normalization to ensure data falls within expected ranges, as can be seen in Figure~\ref{fig:perlin}. For scenarios with multiple features, like the one-hot encoded case, we utilize overlapping Perlin noise to maintain feature dependencies.

\subsection{Related Efforts}
Several works have explored DP for data aggregation and location privacy. PrivStats \cite{popa2011privacy} uses DP and secure computation for private statistical aggregation over distributed datasets. However, it is not specialized for geographical data.

In the location privacy domain, Apple's work on learning iconic scenes \cite{DpApple}, which we are using as a reference for this work, employs local DP and secure aggregation. Mir et al. \cite{6691626} propose a DP framework using spatial decomposition trees for location data aggregation. Aktay et al. \cite{DBLP:journals/corr/abs-2004-04145} apply DP for mobility analysis during COVID-19.

Our work bridges the gap between these two areas, providing a unified DP framework tailored for geographical data aggregation across diverse data types and use cases.

These related efforts can be examined in more detail as follows. 

\subsubsection{Privacy and Accountability for Location-based Aggregate Statistics  \cite{popa2011privacy}}

The focus of this paper is on the issue of location privacy within mobile applications. As apps become increasingly capable of tracking user location, there's a growing risk that private movement data could be compromised. The authors propose a solution called PrivStats. This system is designed to calculate aggregate location statistics (things like how many people are in a certain area), while ensuring that no individual's location privacy is violated. PrivStats safeguards privacy even if the server running the system possesses additional background knowledge about its users.

PrivStats also addresses the problem of accountability. Some users might try to disrupt the system by feeding it false data. The system protects itself through the use of specialized protocols.  These protocols allow anonymous data submission and also include "zero-knowledge proofs" – a way to verify that data is legitimate without revealing additional details. The paper describes how PrivStats was tested and implemented. Experimentation showed that it can efficiently handle a large number of users and responds quickly, showing promise as a practical solution for preserving location privacy in a data-driven world.

\subsubsection{Learning Iconic Scenes with Differential Privacy by Apple \cite{DpApple}}

This is an approach by Apple to detect iconic scenes from photo taken by iOS 17 users. "Iconic Scenes" in this case is defined as the most similar images taken by people from a certain geographical location. For example, 'Near the Golden Gate Bridge' means the most iconic Scene is the bridge and the sunset. Apple's research led to a dashboard as shown in Figure 2. This was created as a method for Apple to provide DP Guarantees to the "Memories" feature in their Photos app. 

\begin{figure*}[h!]
  \includegraphics[width=\textwidth]{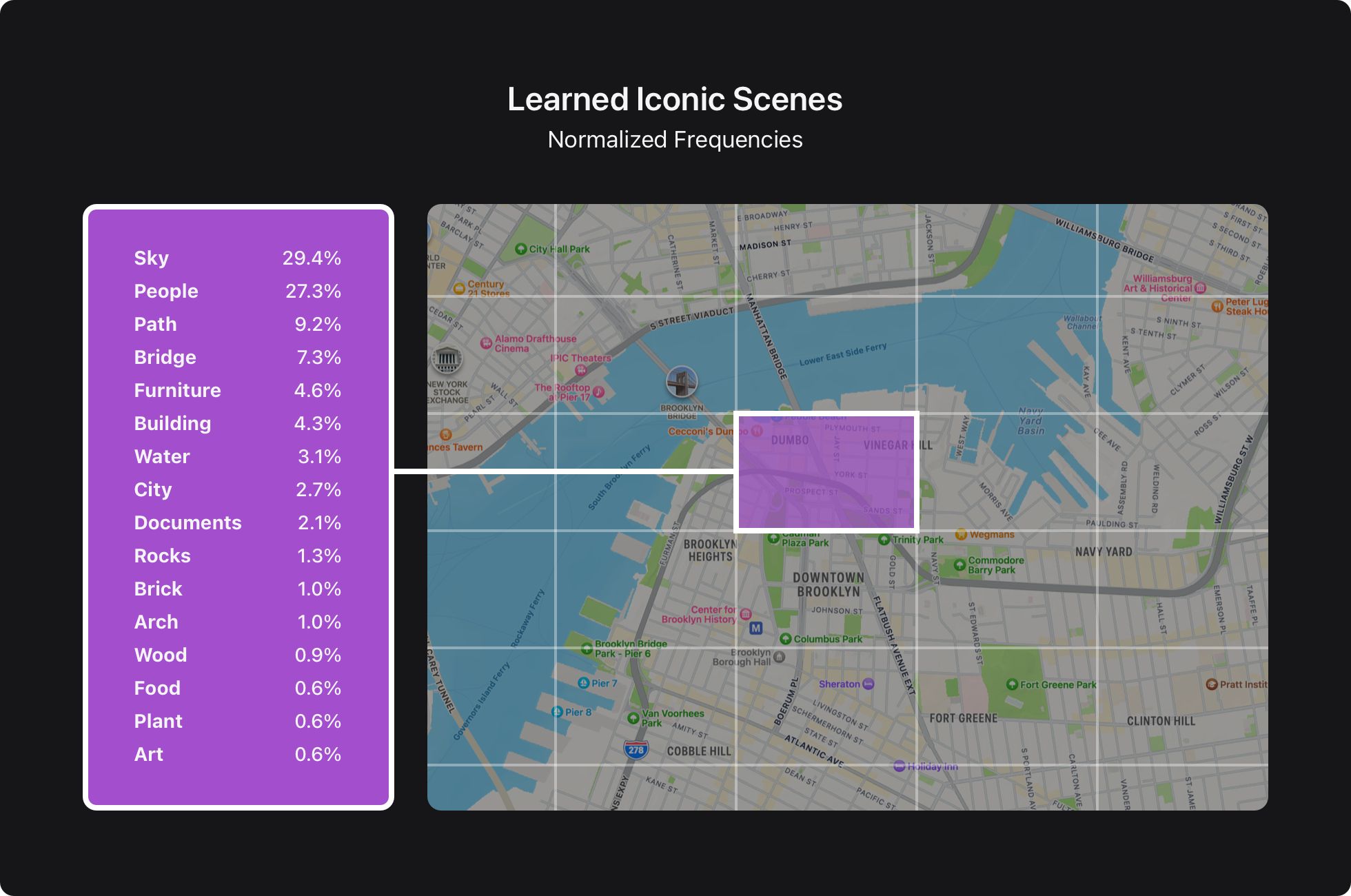}
  \caption{Apple's implementation \cite{DpApple} - Map showing final normalized histograms of top categories for photos taken near the Brooklyn Bridge in New York City.}
  \label{fig:applescenes}
\end{figure*}

To balance privacy with utility, Apple combined local noise addition with secure aggregation. Photos are annotated locally with common categories, such as person, sky, recreation, etc. using a Multi-Task Neural Architecture Model \cite{multitaskmodel}. If the user opts in to the Analytics and Improvements feature, they further processing takes place. Then, a random location-category pair is selected (e.g., Central Park, person) and encoded as a one-hot vector. Noise is introduced by randomly flipping bits for local DP assurance.

As demonstrated in Figure 3, the noisy one-hot vector then gets split into two shares, each encrypted with a separate public key. These are uploaded to a server with two components:  the "leader" (with one private key) and the "helper" (with the other private key). Both components aggregate shares only if a minimum cohort size is reached. Neither component alone can see individual vectors, only DP-assured aggregates.

\begin{figure}[h!]
  \includegraphics[width=0.5\textwidth]{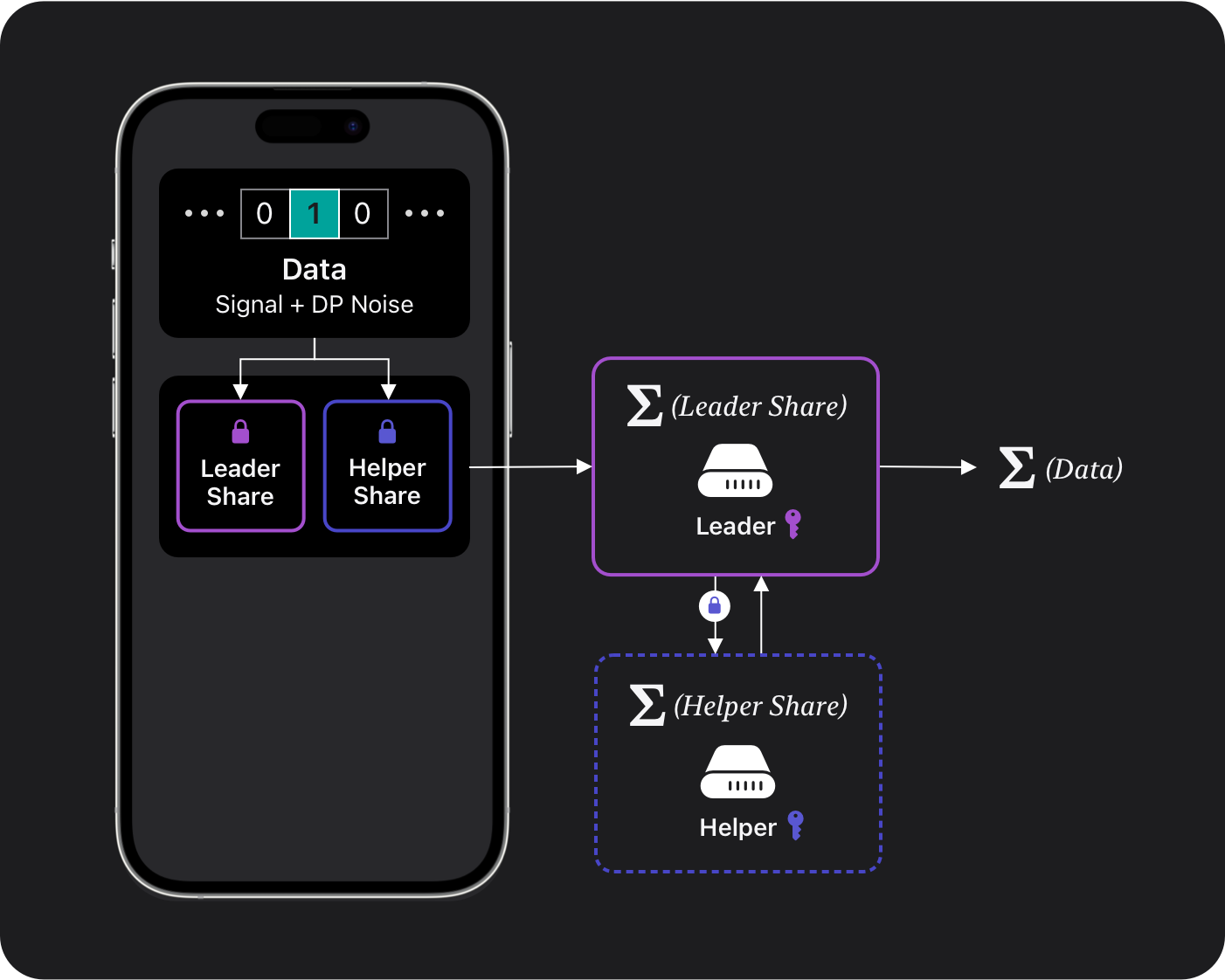}
  \caption{"The secure aggregation protocol that enforces the DP assurance. Each binary vector is split into two shares, and each share is encrypted with a different public key. As long as no single entity gets access to both private keys, nobody can see any individual vectors, only the aggregate, which satisfies the desired DP assurance" \cite{DpApple}}
  \label{fig:leadershare.png}
\end{figure}

The 'helper' and 'leader' combine results for a noisy aggregate of all devices within the cohort. This reveals location-photo category frequencies. To prevent malicious updates without anyone being able to see individual vectors, Apple uses a technique called Prio validation \cite{DBLP:journals/corr/Corrigan-GibbsB17}. Prio validation is a privacy-preserving technique that allows secure computation of aggregate statistics (like sums, means, etc.) over data distributed across multiple parties. It ensures that only valid contributions are included in the final result, preventing malicious actors from corrupting the data without revealing individual user information.

Using this approach, frequencies for 4.5 Million location-category pairs were discovered, with a DP of $\epsilon = 1$, $\delta = 1.5e-7$. These frequencies are then utilized to improve photo selection within Apple products.

\section{Preliminaries}

As our work is centered around DP, it will be important for this to be defined. Differential privacy (DP) is a rigorous mathematical framework that provides strong privacy guarantees for data analysis and processing. It ensures that the output of a computation does not reveal too much information about any individual data point in the input dataset. DP achieves this by introducing carefully calibrated noise into the computation, making it difficult to distinguish between the presence or absence of any particular record. This privacy guarantee is parameterized by a privacy budget, denoted as $\epsilon$, which controls the trade-off between privacy and utility. A smaller $\epsilon$ value provides stronger privacy guarantees but may result in reduced accuracy of the computed results.

\textit{Definition 1 ($\epsilon$-DP)} A randomized mechanism $\mathcal{M}$ satisfies $\epsilon$-DP if for any two neighboring datasets $D$ and $D'$, differing by at most one record, and any subset $S \subseteq Range(\mathcal{M})$:

\begin{equation*}
\Pr[\mathcal{M}(D) \in S] \leq e^\epsilon \cdot \Pr[\mathcal{M}(D') \in S]
\end{equation*}

In our framework, we employ three local DP mechanisms to achieve privacy-preserving data aggregation: randomized response, the exponential mechanism, and the Gaussian mechanism. Each mechanism offers unique properties and is suitable for different data types and aggregation scenarios. 

We will now proceed to explore each mechanism in detail. 

\subsection{Randomized Response}
Randomized response is a classic DP technique that allows individuals to locally perturb their sensitive attributes before sharing them with an untrusted aggregator. It provides plausible deniability to the respondents, as the true value of their attribute is obscured by the randomization process. Randomized response is particularly well-suited for binary or categorical attributes.

In the case of binary attributes, randomized response achieves $\epsilon$-DP by flipping the true attribute value with a specific probability determined by the privacy budget $\epsilon$. The flipping probability is given by $\frac{1}{1+e^\epsilon}$. By tuning the value of $\epsilon$, one can control the level of privacy provided by the randomized response mechanism.

\subsection{Exponential Mechanism}
The exponential mechanism is a versatile DP technique that privately selects an output from a set of possible choices based on a given utility function. It is particularly useful when the output space is large or not naturally ordered. The exponential mechanism assigns higher probabilities to outputs that have higher utility scores, while ensuring that the selection process satisfies $\epsilon$-DP.

Formally, for an output space $\mathcal{O}$, a dataset $D$, a utility function $u : \mathcal{O} \times D \rightarrow \mathbb{R}$, and a privacy budget $\epsilon$, the exponential mechanism $\mathcal{M}_E$ selects an output $r \in \mathcal{O}$ with probability proportional to $\exp(\frac{\epsilon \cdot u(r, D)}{2\Delta u})$, where $\Delta u$ is the sensitivity of the utility function.

\subsection{Gaussian Mechanism}
The Gaussian mechanism is a DP technique that adds carefully calibrated Gaussian noise to the true output of a function to achieve privacy protection. It is commonly used for numerical queries or functions with continuous output ranges. The magnitude of the added noise depends on the sensitivity of the function and the desired level of privacy.

For a function $f : D \rightarrow \mathbb{R}^d$ with $l_2$-sensitivity $\Delta_2 f$, the Gaussian mechanism $\mathcal{M}_G$ produces a noisy output by adding Gaussian noise with zero mean and variance $\sigma^2$ to the true output $f(D)$. The standard deviation $\sigma$ is determined based on the privacy budget $\epsilon$ and the target $\delta$, which represents the probability of privacy failure. To achieve $(epsilon, delta)$-DP, the Gaussian mechanism requires $\sigma \geq \frac{\Delta_2 f \sqrt{2\ln(1.25/\delta)}}{\epsilon}$.

These three local DP mechanisms form the foundation of our privacy-preserving data aggregation framework. By carefully selecting and applying these mechanisms based on the specific data types and aggregation scenarios, we ensure that the privacy of individual records is protected while still allowing for meaningful insights to be derived from the aggregated data.

\section{Implementation Details}

\begin{figure*}[h!]
  \includegraphics[width=\textwidth]{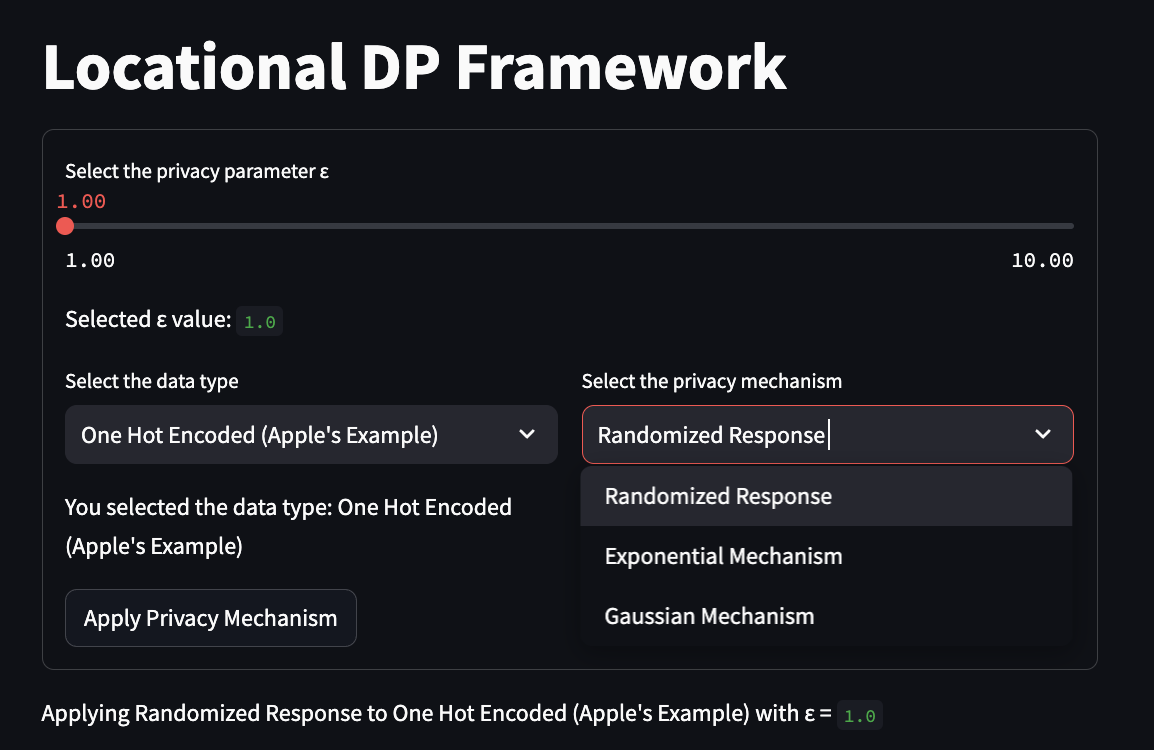}
  \caption{Simulator Front-end: UI of Streamlit app to allow users to select $\epsilon$ value, data type and privacy mechanism to apply to preset map region.}
  \label{fig:simul}
\end{figure*}

\begin{figure*}[h!]
  \includegraphics[width=1.05\textwidth]{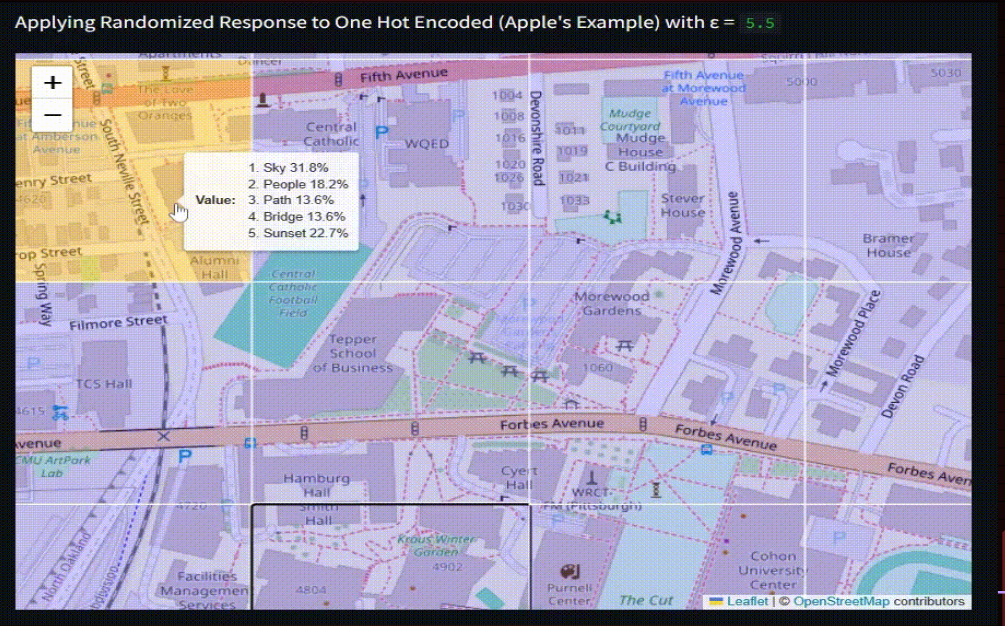}
  \caption{Simulator Result: Map showing final normalized histograms of top categories for photos taken near the CMU Campus in Pittsburgh, PA.}
  \label{fig:demomap}
\end{figure*}

\begin{figure*}[h!]
  \includegraphics[width=\textwidth]{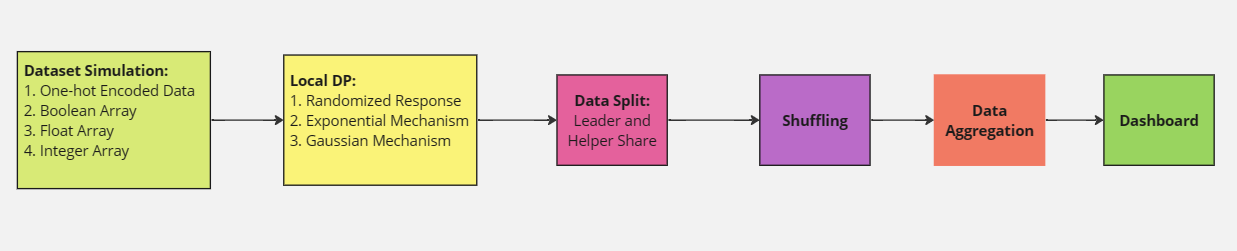}
  \caption{Flowchart for Our Locational DP Framework Implementation}
  \label{fig:flow}
\end{figure*}

Our implementation aims to preserve the compute efficiency of aggregate results while providing DP guarantees. Inspired by Apple's work, we implement shuffling for additional privacy amplification but do not fully replicate their secure aggregation framework due to time constraints.

We prioritize utility and measure the difference between results with no DP (i.e., $\epsilon = \infty$) and with DP applied. Privacy budgets are tracked and recorded throughout the aggregation process using the Opacus library.

Our framework is designed to be modular, allowing easy integration of additional local DP mechanisms and datasets as needed. Our framework also allows users to experiment with different noise mechanisms such as exponential, Gaussian, or randomized response. We provide a synthetic data generator for different data types such as integer (rankings), boolean (contagion tracking), float (income), and one-hot encoding (Apple's implementation) based on a given distribution. This is also designed in such a way that it is extensible onto other data types.

We also provide an accessible front-end, which allows users to simulate and test the mechanisms with different parameters such as Gaussian, exponential or randomized response. With an option to select an $\epsilon$ value, users can simulate and understand the usability versus privacy trade-off of different mechanisms and $\epsilon$ configurations. This can be seen in Figure 4 and Figure 5. The code for our implementation can be found at 
\url{https://github.com/Privacy-Engineering-CMU/location-dp-framework}. As for the general flow of our implementation, that is illustrated in Figure 6. 

\label{fig:simul}

\section{Results}

\begin{figure*}[h!]
  \includegraphics[width=\textwidth]{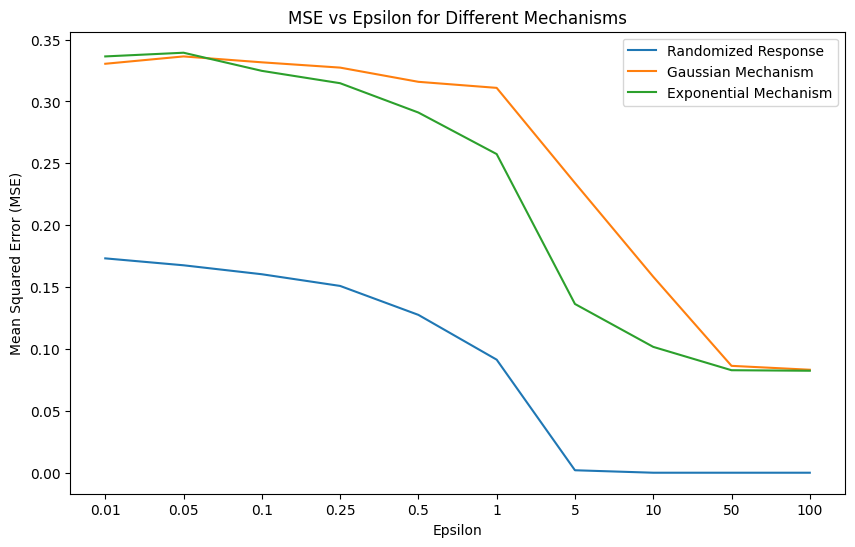}
  \caption{Mean Squared Error (MSE) averaged across scenarios for varying $\epsilon$ values, comparing the performance of the three local DP mechanisms: randomized response, exponential mechanism, and Gaussian mechanism. The graph demonstrates an exponential decrease in MSE as $\epsilon$ increases for all three mechanisms, indicating that as more privacy is sacrificed (i.e., larger $\epsilon$ values), the utility of the aggregated data improves.}
  \label{fig:results}
\end{figure*}
We will now evaluate our framework on the four simulated datasets described in Section 2.2, measuring utility in terms of Mean Squared Error (MSE) between the non-private and DP results. Figure 6 shows these results. 

Across all datasets and mechanisms, we observe the expected trend of increasing $\epsilon$ leading to reduced error and improved utility at the cost of weaker privacy guarantees. The exponential and Gaussian mechanisms generally outperform randomized response in terms of utility, with the Gaussian mechanism providing the best performance for numerical data.

\section{Discussion}

This unified DP framework marks our effort towards enabling privacy-preserving geographical data aggregation for a wide range of applications. Our approach effectively addresses the challenges associated with diverse data types and aggregation scenarios, offering a flexible and adaptable solution. The modular design allows for the integration of various local DP mechanisms, such as randomized response, the exponential mechanism, and the Gaussian mechanism, tailoring the framework to the specific privacy requirements and data characteristics of different use cases.

One of the key strengths of our framework lies in its ability to strike a balance between privacy protection and data utility. By employing techniques like shuffling and privacy budget tracking, we ensure robust privacy guarantees while maintaining the usefulness of the aggregated data for analysis and decision-making purposes. This balance is crucial for the practical adoption and deployment of privacy-preserving technologies in real-world settings, where both privacy and utility are of paramount importance.

However, it is important to recognize that our approach is not without limitations. The inherent trade-off between privacy and utility in DP mechanisms means that there may be some impact on the accuracy and granularity of the insights derived from the aggregated data. Striking the optimal balance requires careful consideration and tuning of privacy parameters, as well as efficient allocation of the privacy budget across multiple aggregation steps. Furthermore, the computational overhead introduced by local DP mechanisms can present scalability challenges, particularly when dealing with large-scale datasets or real-time aggregation scenarios, necessitating further research into optimization techniques and efficient implementations.

\subsection{Laws \& Regulations}

The implementation of DP in the aggregation of geographical data must adhere to several legal and regulatory frameworks to ensure the protection of individual privacy and compliance with data protection laws. This section outlines a few primary laws and regulations that govern our DP framework.

\subsubsection{General Data Protection Regulation (GDPR)}

The GDPR \cite{http://data.europa.eu/eli/reg/2016/679/oj} is a critical regulatory framework for companies operating in the European Union or dealing with EU citizens' data. Our DP framework aligns with GDPR's principles by minimizing the amount of personal data processed, especially sensitive locational data, and incorporating robust measures to ensure data confidentiality and integrity. Under GDPR, the processing of personal data using DP is considered a form of pseudonymization, which is encouraged as a risk reduction technique.

\subsubsection{California Consumer Privacy Act (CCPA)}

For users in California, the CCPA provides rights similar to GDPR, including the right to know about the personal information a business collects and the purpose for its use \cite{CCPA2020}. Our framework's use of DP ensures that individual data cannot be reconstructed or linked back to any user without their explicit consent, supporting compliance with CCPA requirements.

\subsubsection{Health Insurance Portability and Accountability Act (HIPAA)
}

While HIPAA primarily applies to health information, our framework's principles for securing data can similarly protect sensitive health-related geographic data aggregations, such as those used during the COVID-19 pandemic \cite{HIPAA1996}. By implementing local DP within an appropriate epsilon value, we ensure that any health data used cannot be traced back to an individual, thus maintaining compliance with HIPAA's Privacy Rule.

\section{Conclusion \& Future Work}
In this paper, we presented a unified location DP framework for privately aggregating diverse data types over geographical regions. Our framework combines local DP mechanisms with shuffling and privacy budget tracking to provide formal privacy guarantees while enabling utility for a range of applications.

Future work includes integrating secure aggregation protocols, extending the framework to additional data types and mechanisms, and evaluating on real-world datasets. We also plan to package our framework into an open-source library and release it to facilitate private geographical data analysis in the wider research community.

\section{Ethical Considerations and Limitations}
While our framework provides strong DP guarantees, it is important to acknowledge that DP is a statistical privacy notion and does not protect against all potential attacks or inferences. Careful privacy budget management and thoughtful selection of $\epsilon$ and $\delta$ values are crucial for ensuring meaningful privacy protection.

Additionally, our simulated datasets, while designed to capture key characteristics of real-world data, may not fully reflect the complexities and biases present in actual geographical datasets. Applying our framework to real-world data may require additional considerations around fairness, representativeness, and potential disparate impacts.

\newpage

\bibliography{custom}
\bibliographystyle{acl_natbib}

\end{document}